\documentclass[final]{cvpr}
\usepackage{times}
\usepackage{epsfig}
\usepackage{graphicx}
\usepackage{amsmath}
\usepackage{amssymb}

\usepackage[pagebackref=true,breaklinks=true,colorlinks,bookmarks=false]{hyperref}

\usepackage{url}            
\makeatletter
\g@addto@macro{\UrlBreaks}{\UrlOrds}
\makeatother
\usepackage{booktabs}       
\usepackage{amsfonts}       
\usepackage{nicefrac}       
\usepackage{microtype}      

\usepackage{comment}
\usepackage{color}

\usepackage{array}
\usepackage{cases}
\usepackage{booktabs}
\usepackage{multirow}
\usepackage{colortbl}
\usepackage[linesnumbered, ruled]{algorithm2e}
\usepackage{algpseudocode}

\newcommand{\eq}{Eq.}

\newcommand{\proposed}{UTAD}
\definecolor{mygray}{gray}{.9}

\def  \LL      {\mathcal{L}}                 
\def  \DD      {\mathcal{D}}                 
\def  \MM      {\mathcal{M}}                 

\begin{document}

\title{Unsupervised Two-Stage Anomaly Detection}


\author{Yunfei~Liu\quad\quad~Chaoqun~Zhuang\quad\quad~Feng~Lu\thanks{ Corresponding author. }\\
State Key Laboratory of Virtual Reality Technology and Systems, \\
School of Computer Science and Engineering, Beihang University, Beijing, China  \\
\small{\texttt{\{lyunfei, zhuangchaoqun, lufeng\}@buaa.edu.cn}}
}

\maketitle

\begin{abstract}
   
    Anomaly detection from a single image is challenging since anomaly data is always rare and can be with highly unpredictable types. With only anomaly-free data available, most existing methods train an AutoEncoder to reconstruct the input image and find the difference between the input and output to identify the anomalous region. However, such methods face a potential problem – a coarse reconstruction generates extra image differences while a high-fidelity one may draw in the anomaly (Fig.1). In this paper, we solve this contradiction by proposing a two-stage approach, which generates high-fidelity yet anomaly-free reconstructions. Our Unsupervised Two-stage Anomaly Detection (\proposed) relies on two technical components, namely the Impression Extractor (IE-Net) and the Expert-Net. The IE-Net and Expert-Net accomplish the two-stage anomaly-free image reconstruction task
    while they also generate intuitive intermediate results, making the whole \proposed~ interpretable. Extensive experiments show that our method outperforms state-of-the-arts on four anomaly detection datasets with different types of real-world objects and textures.

\end{abstract}


\section{Introduction}

Most objects or textures in nature have certain shapes and properties, especially human-made ones. 
Anomaly detection identifies rare items, events, or observations that raise suspicions by differing significantly from the majority~\cite{A:arthur2016outlier}.
It will translate to many computer vision tasks, such as detect defective product parts~\cite{A:bergmann2019mvtec}, segment lesions in retinopathy images~\cite{A:niu2019pathological}, and locate intruders in surveillance~\cite{A:nguyen2018weakly}, \etc.
Anomaly detection is challenging since most deep learning-based methods require balanced positive data and negative data for training. However, abnormal data is always limited, and hard or even cannot be obtained in terms of their amount and types. 

To tackle the lack of abnormal data, many methods assume the novel things that occurred on the \textit{category} or \textit{image-level} are anomalies. They detect if a new input is out-of-distribution when compared with the training data. These methods can be referred to as one-class-classification or outlier detection~\cite{A:andrews2016transfer,A:napoletano2018anomaly,A:ruff2020deep,A:tax2004support}. However, 
this setting is not suitable for anomaly detection, which pays more attention on the images within the same category. In this case, anomalies often exist in small areas in the object or image (\eg, crack on the surface, missing small parts, \etc).

\begin{figure*} 
	\begin{center}
		\includegraphics[width=0.95\linewidth]{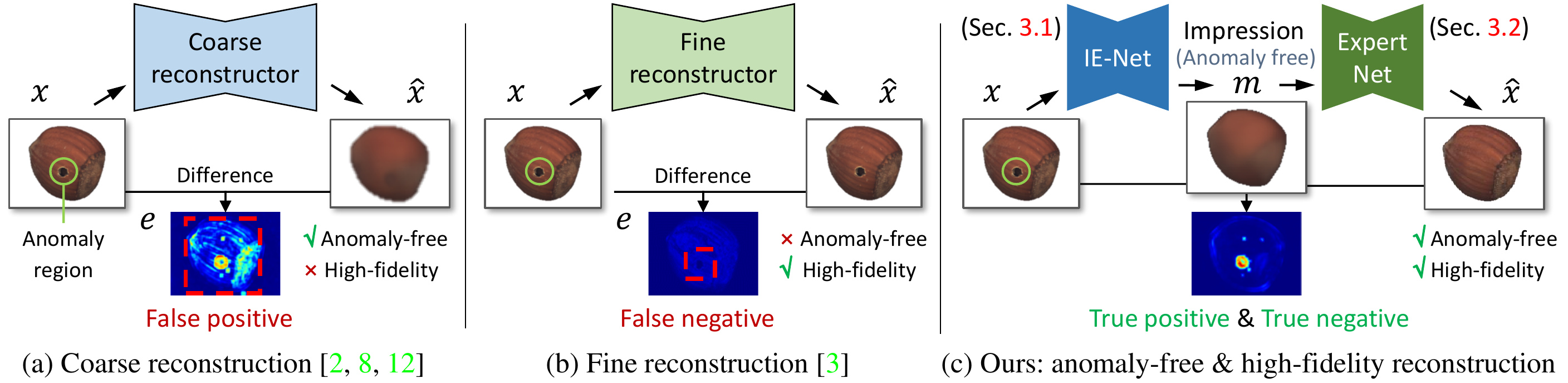} 
	\end{center}
	\vspace{-0.3cm}
	\caption{
		Comparisons of different anomaly detection methods. 
		Many one-stage methods reconstruct the structure of the input but without details (A)~\cite{A:akcay2018ganomaly,A:baur2018deep,A:bergmann2018improving} or high-fidelity reconstruction yet draw in anomaly regions (B)~\cite{A:akccay2019skip}.
		(c) We tackle the contradiction by extracting the anomaly-free structure ($m$) and adding high-fidelity details in two stages.
	}
	\vspace{-0.4cm}
	\label{fig:diff}
\end{figure*}

Recently, many approaches tackle this challenging problem with AutoEncoders~\cite{A:baur2018deep,A:bergmann2018improving,A:carrera2016defect,A:gong2019memorizing} and GAN based methods~\cite{A:akcay2018ganomaly,A:akccay2019skip,A:schlegl2017unsupervised}, which assume the network cannot reconstruct the unseen regions, then find the input and its difference with the output to identify the anomaly. 
Although such methods are able to distinguish novel classes from old ones (\eg, find the novel class in MNIST~\cite{B:lecun1998MNIST} or CIFAR-10~\cite{B:krizhevsky2009CIFAR10}, \etc), these methods face a potential contradiction. 
As shown in Fig.\ref{fig:diff} (a), coarse reconstructions generate extra image difference, which interrupts the anomaly detection. To reconstruct more high-fidelity details, many methods\cite{A:akccay2019skip,A:ronneberger2015unet} equip with more complex architectures or more trainable parameters. However, the output may draw the anomaly in and loses the anomaly-related difference.

In this paper, we tackle this contradiction with two steps to ensure anomaly-free and high-fidelity, respectively.
More specifically, we introduce a mediate state, namely impression $m$, which is the anomaly-free reconstruction of the input. 
Then we design a two stage framework to generate both anomaly-free and high-fidelity reconstruction for anomaly detection.
As illustrated in Fig.~\ref{fig:diff} (c), our idea is to generate the anomaly-free and high-fidelity $\hat{x}$ through two stages.
Consequently, we introduce the Unsupervised Two-stage Anomaly Detection (\proposed) framework, which contains two key components: the Impression Extractor (IE-Net) and the Expert-Net. The IE-Net is used for the first-stage reconstruction, which generates the anomaly-free $m$ for the given input $x$ (Sec.~\ref{sec:IE}). The Expert-Net then restores details on $m$ for the high-fidelity $\hat{x}$ (Sec.~\ref{sec:EN}). Finally, we use Perceptual Measurement (\textit{PM}) to better detect anomaly-related differences among the input $x$, impression $m$ and reconstructed $\hat{x}$ (Sec.~\ref{sec:PM}).

In summary, our main contributions are:
\begin{itemize}
    \item We propose a novel Unsupervised Two-stage Anomaly Detection (\proposed) framework that enables both anomaly-free and high-fidelity reconstruction for anomaly detection from a single image. The method requires no anomaly sample for training. 
    \item We propose the new concept of anomaly-free reconstruction (namely impression) for anomaly detection. We design an IE-Net and Expert-Net to extract and utilize impression for anomaly-free and high-fidelity reconstructions, respectively. 
    \item The proposed \proposed~ framework outperforms state-of-the-arts on four anomaly detection datasets with different types of real-world objects and textures. 
\end{itemize}

\begin{figure*}
	\begin{center}
		\includegraphics[width=0.95\linewidth]{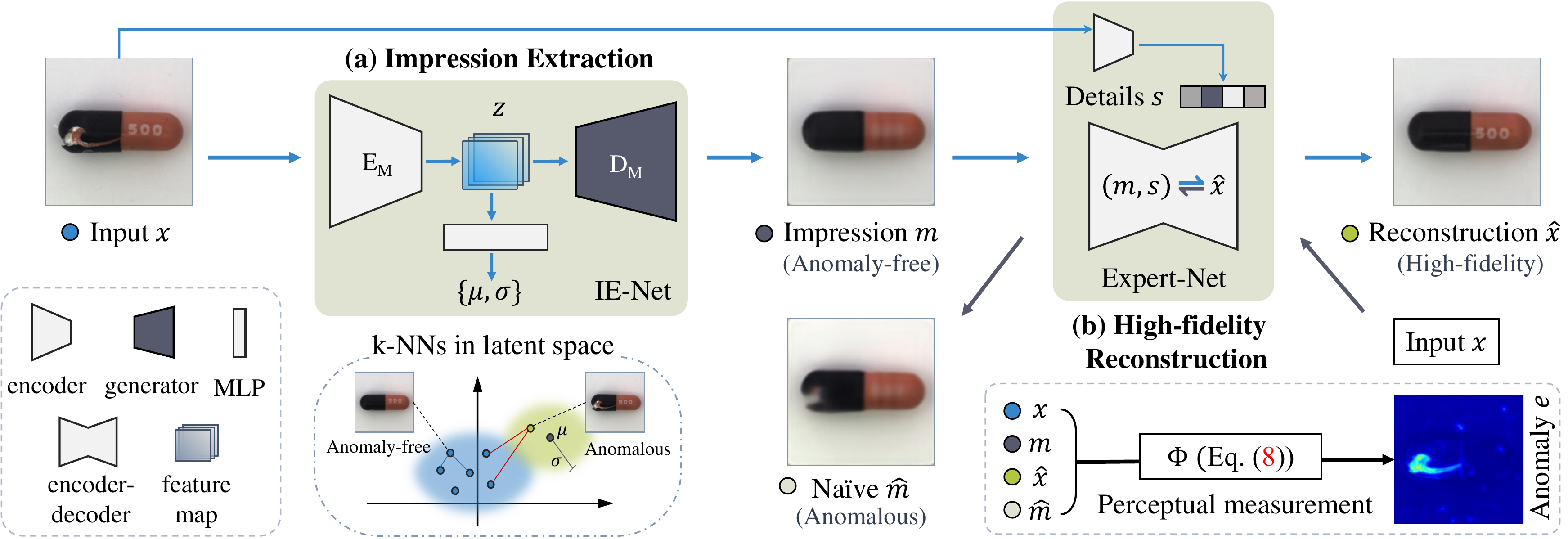} 
	\end{center}
	\vspace{-0.2cm}
	\caption{Overview of the proposed method. Our framework mainly contains two components: \textit{IE-Net} and \textit{Expert-Net}. 
	(a) IE-Net tries to generate the anomaly-free reconstruction (\ie, impression $m$).
	(b) The \textit{Expert-Net} aims to learn an invertible mapping to manipulate the high-fidelity details between impression $m$ and input $x$.
	Finally, the anomaly map $e$ is calculated through \textit{Perceptual Measurement} $\Phi$ by leveraging the difference among $x, \hat{x}, m, \text{and}~\hat{m}$.
	}
	\label{fig:overview}
	\vspace{-0.3cm}
\end{figure*}

\section{Related Works}

\subsection{Unsupervised anomaly detection}

Many AutoEncoder-based methods are proposed for unsupervised anomaly detection.
Carrera \etal~\cite{A:carrera2016defect} train an AutoEncoder and make it over-fit on the anomalous-free images, then use the magnitude of reconstruction loss (\ie, MSE loss) on test images to determine anomalous regions. Based on this, Bergmann \etal~\cite{A:bergmann2018improving} propose to replace per-pixel MSE loss with structure similarity loss~\cite{B:wang2004SSIM}. 
Baur~\etal~\cite{A:baur2018deep} propose to use a variational auto-encoder (VAE) instead of the auto-encoder. There are also many GAN~\cite{B:radford2015DCGAN} based methods for anomaly detection. AnoGAN~\cite{A:schlegl2017unsupervised} is the first to use a generator for unsupervised anomaly detection on retinopathy images. Inspired by AnoGAN, GANormaly~\cite{A:akcay2018ganomaly} adds an encoder for mapping images to latent space, and greatly decreases the inference time. Fast-AnoGAN~\cite{A:schlegl2019fastAnnoGAN} trains a WGAN~\cite{B:arjovsky2017WGAN} and an additional encoder with two stages to boost the performance. 
Berg \etal~\cite{A:berg2019unsupervised} propose to combine progressive growing GAN~\cite{B:karras2017progressive} and ClusterGAN~\cite{A:mukherjee2019clustergan} together to reconstruct high-resolution images for anomaly detection.
Skip-GANomaly~\cite{A:akccay2019skip} replaces the auto-encoder with U-net~\cite{A:ronneberger2015unet} and further gets better reconstructions. 
However, such methods face a potential problem of whether making a coarse reconstruction or a high-fidelity one – the former generates extra image differences while the latter may draw the anomaly in and lose the anomaly-related difference.

Based on the uncertainty learning, Bergmann~\etal\cite{A:bergmann2020uninformed} propose Uniformed Students, which use the teacher-students for unsupervised anomaly detection. However, the learning process is kind of verbose and with limited interpretation. 
Venkataramanan~\etal~\cite{A:venkataramanan2020attention} propose to use an attention mechanism, and use the activated feature for anomaly detection. There are also other methods that use additional supervision~\cite{A:zhou2020encoding} or memory bank~\cite{A:gong2019memorizing} for anomaly detection. 

\subsection{One class classification}
One class classification (\ie, outlier detection) is concerned with distinguishing out-of-detection samples relative to the training set. 
SVDD~\cite{A:tax2004support} tries to map all the normal training data into predefined kernel space and takes the sample that outside the learned distribution as anomaly.
Andrews \etal~\cite{A:andrews2016transfer} use the different layer features from the pre-trained VGG network and model the anomaly-free images with a $v$-SVM.  
There are many methods that equip this idea with patched distribution clustering~\cite{A:napoletano2018anomaly} or extend it to a semi-supervised scenario~\cite{A:ruff2020deep}.
There is a major difference between the one-class classification and anomaly detection since one class classification detects images from a different category, while anomaly detection aims to detect and locate the difference between images that are from the same class.

\subsection{Unsupervised Representation learning}
Learning a good representation of an image is a long-standing problem of computer vision. 
One branch of research suggests training the encoder by learning with
a \textit{pretext task} (\eg, predicting relative patch location~\cite{R:doersch2015context_prediction}, solving a jigsaw puzzle~\cite{R:noroozi2016jigsaw}, colorizing images~\cite{R:zhang2016colorization}, counting objects~\cite{R:noroozi2017count}, and predicting rotations~\cite{R:gidaris2018rotations}.). 
Another branch of research suggests extracting the unique information of the sample so that the sample can be distinguished easily for the downstream tasks. Based on this assumption, many mutual information-based methods are proposed, like f-GAN~\cite{R:nowozin2016fGAN}, info-GAN~\cite{R:chen2016infogan}, info-NCE~\cite{R:tschannen2020mutual_InfoNCE}, and AMDIM~\cite{R:bachman2019learning_AMDIM}, \etc. In this paper, we are inspired by the mutual information and propose to learn the distinctive and anomaly-free features for the anomaly detection task.

\section{Unsupervised Two-stage Anomaly Detection}

Here we introduce the proposed Unsupervised Two-stage Anomaly Detection (\proposed) for unsupervised anomaly detection. Given a training set $\DD = \{x_1, x_2, \dots, x_N\}$ of anomaly-free images, our goal is to learn two different networks that serve two stages for anomaly detection without any anomalous samples.
As shown in Fig.~\ref{fig:overview}, \proposed~ is constructed by 
\begin{itemize}
    \item \textit{Impression Extractor Network} (IE-Net) extracts the anomaly-free impression $m$ on the input $x$.
    \item \textit{Expert-Net} is an invertible mapping, which generates details on $m$ to produce the high-fidelity reconstruction $\hat{x}$. Meanwhile, for a better performance and interpretation, it also produces an intermediate na\"ive impression $\hat{m}$ by mimicking the appearance of the output of IE-Net.
\end{itemize}

\subsection{IE-Net for anomaly-free reconstruction} \label{sec:IE}

IE-Net aims to reconstruct the anomaly-free impression $m$ for anomaly detection. 
The output of IE-Net is used as the input of Expert Net, which will be introduced in Sec.~\ref{sec:EN}.

\begin{figure}
	\begin{center}
		\includegraphics[width=0.9\linewidth]{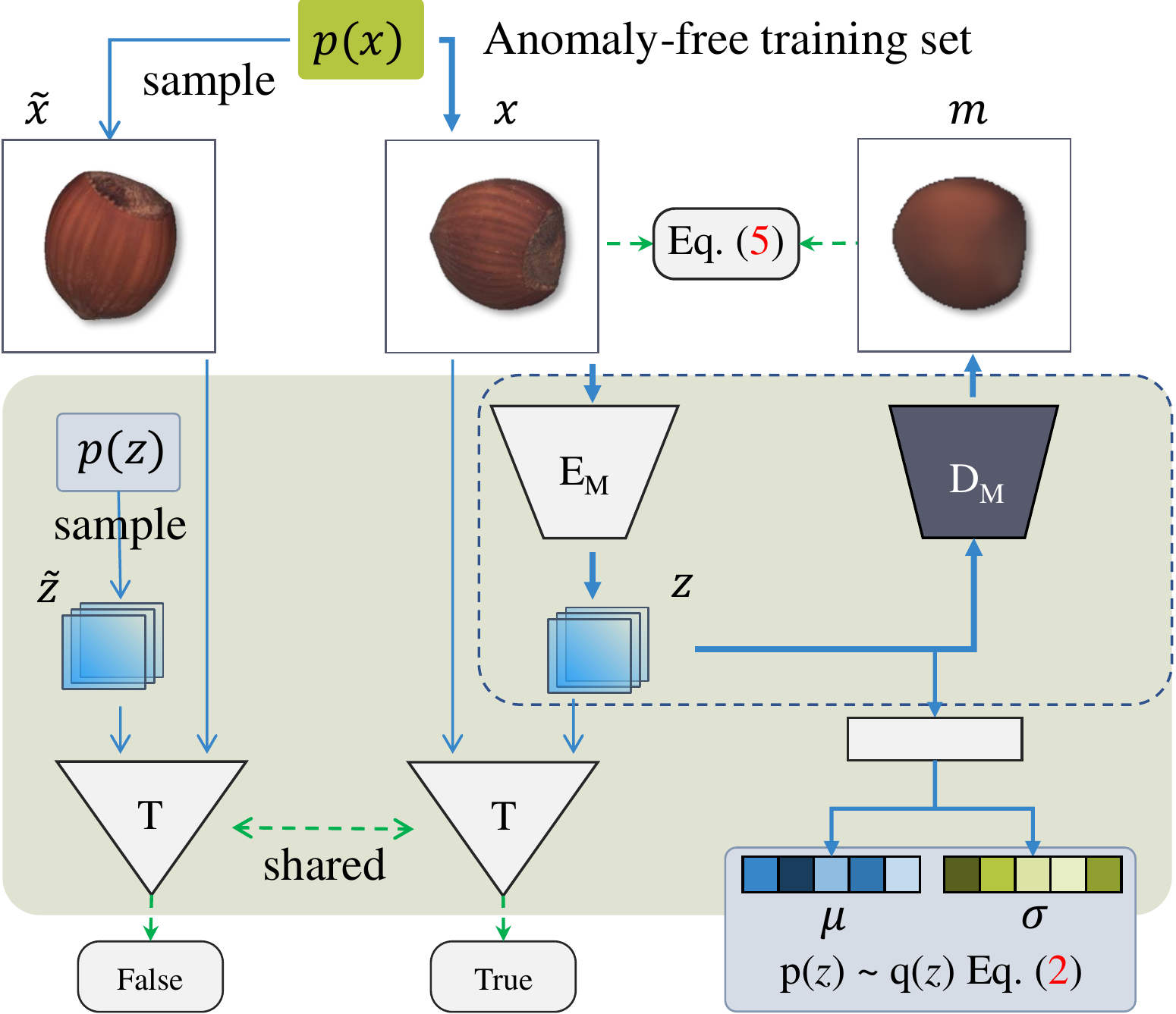} 
	\end{center}
	\caption{Architecture of the IE-Net, which is constructed by an encoder $E_M$, a discriminator $T$, and a decoder $D_M$. When training is done, the components in the dashed box are used for impression extraction.
	}
	\label{fig:info_cae}
\end{figure}

\noindent\textbf{Training phase.}
The IE-Net is trained on the anomaly-free dataset. The architecture of IE-Net is illustrated in Fig.~\ref{fig:info_cae}. IE-Net is constructed by an encoder $E_M$, a discriminator $T$ and a decoder $D_M$. These modules are jointly optimized by minimizing objective Eq.~\eqref{eq:ie_total_loss}.

\noindent\textbf{Inference phase.}
When IE-Net is trained on the anomaly-free dataset, its encoder $E_M$ and $D_M$ are used for generating anomaly-free impression $m$ for the input image, which may contain anomalous regions. 

Since it is not trivial to control the reconstruction ability for anomaly detection, as illustrated in Fig.~\ref{fig:diff}, we propose to extract the input image with anomaly-free {impression} $m$ on the basis of mutual information theory. 
Specifically, we equip IE-Net with mutual information mainly because 
1) mutual information is used to describe the distinctive features of the input image~\cite{R:tschannen2020mutual_InfoNCE} 
and 2) only anomaly-free samples are used for training. 
Therefore, the anomaly-free and distinctive features are suitable for reconstructing the impression.

As illustrated in Fig.~\ref{fig:info_cae}, 
the encoder $E_M$ aims to extract the distinctive feature $z$ of the image $x$ by maximizing the mutual information $I(\DD, Z)$, where $z \in Z$, $Z$ is the set of latent codes. $I(\DD, Z)$ is defined as follow
\begin{equation} \label{eq:mutual_info}
    I(\DD, Z) = \int\int p(z|x)p(x)\log\frac{p(z|x)}{p(z)}dxdz,
\end{equation}
where $p(x)$ is the distribution of $\DD$.
Following the assumptions in VAE~\cite{B:kingma2013VAE}, the distribution $p(z)$ is required to follow the Gaussian distribution $q(z)$, which is implemented by Kullback-Leibler (KL) divergence
\begin{equation} \label{eq:kl_div}
    KL(p(z)\parallel q(z)) = \int p(z)\log\frac{p(z)}{q(z)}dz.
\end{equation}
As with maximizing the mutual information, the loss function of the encoder $E_M(x)$ becomes
\begin{equation}\label{eq:loss_encoder}
    \LL_M^e = -I(\DD, Z) + \lambda KL(p(z)\parallel q(z))\},
\end{equation}
where $\lambda$ is the hyper-parameter for balancing these two terms in the objective function.
To optimize \eq~\eqref{eq:mutual_info}, we follow f-GAN~\cite{R:nowozin2016fGAN} and convert the maximizing process of mutual information to discriminate (with the discriminator $T(x, z)$) the difference between positive samples $(x, z)$ and negative samples $(\tilde{x}, \tilde{z})$, where $\tilde{x}\in \{\DD\setminus x\}$ and $\tilde{z}$ is a sample from $p(z)$. 
For the convenience of implementation, $\tilde{x}$ is a batch-shuffled version of $x$ and $\tilde{z} \in N(\mu, \sigma^2)$, where $\mu$ and $\sigma$ are the mean and variance of $p(z)$. The $\mu$ and $\sigma$ are estimated by a MLP.
Thus, the loss function of the discriminator $T(x, z)$ is 
\begin{equation}
    \LL_M^t = -\log(T(x, E_M(x))) - \log(1 - T(\tilde{x}, \tilde{z})).
\end{equation}
Thus, the Eq.~\eqref{eq:loss_encoder} became $\LL_M^e = \LL_M^t + \lambda KL(p(z)\parallel q(z))$.

In the next, the decoder $D_M(z)$ tries to project latent code $z$ into RGB-space (\ie, impression $m$) and makes $m$ similar to the input image $x$. The objective function of $D_M(z)$
\begin{equation} \label{eq:impre_rec}
    \LL_M^d =  |D_M(E_M(x)) - x|.
\end{equation}

Therefore, the total loss of IE-Net is
\begin{equation} \label{eq:ie_total_loss}
    \LL_{IE} = \LL_M^e + \lambda_1 \LL_M^d,
\end{equation}
here we add the term $\lambda_1$ to make the objective more flexible.

\subsection{Expert-Net for high-fidelity reconstruction} \label{sec:EN}

\begin{figure}
    \begin{center}
    	\includegraphics[width=0.9\linewidth]{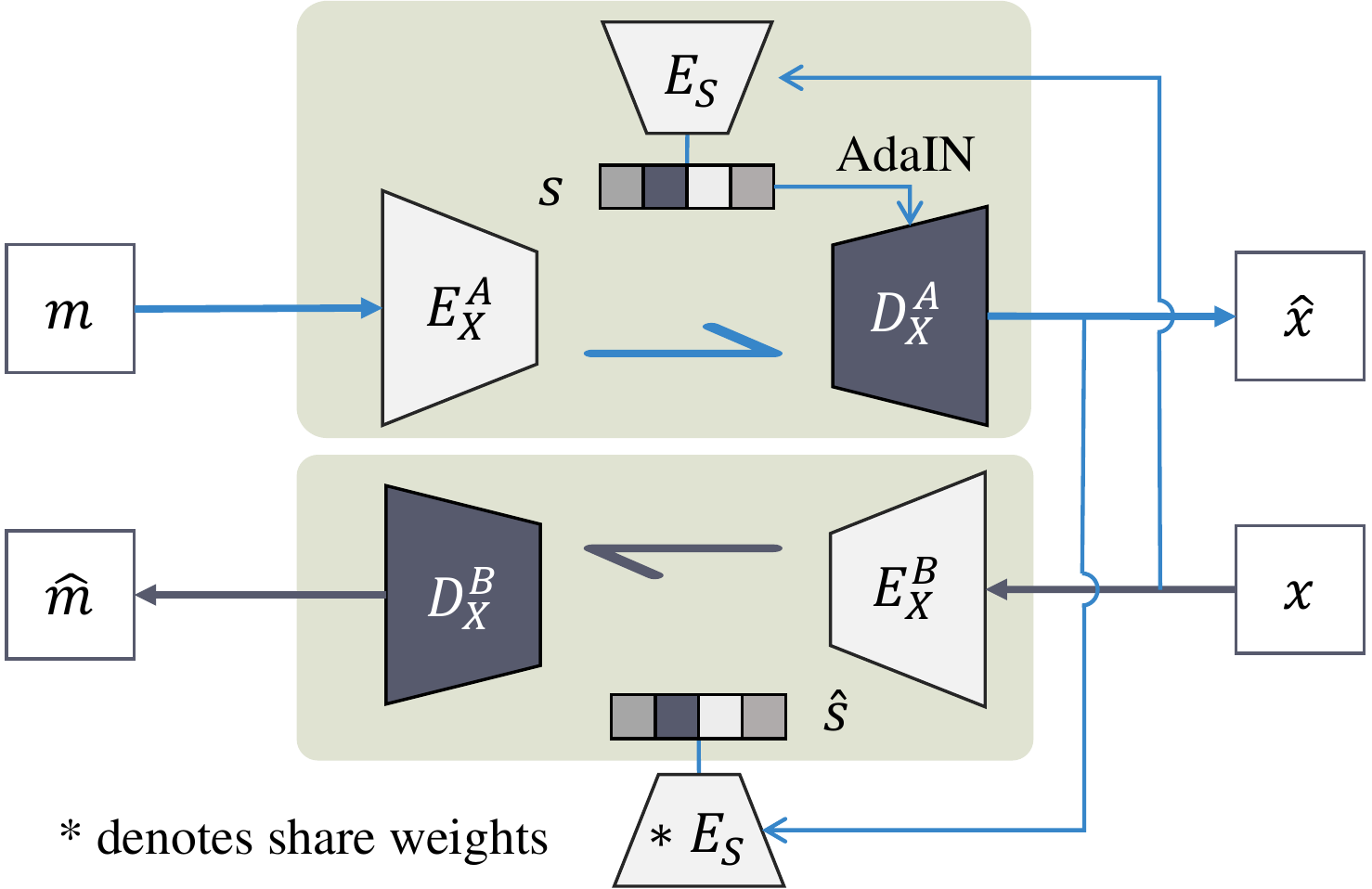} 
    \end{center}
	\vspace{-0.2cm}
    \caption{Architecture of the Expert-Net, which contains a detail extractor $E_S$, two encoders, and two decoders. The invertible mapping between impressions and images is learned.}
    \label{fig:expert_pix2pix}
	\vspace{-0.4cm}
\end{figure}

Through IE-Net, we can generate the anomaly-free impression $m$ for the input image $x$.
We find the $m$ is often with blurry texture or color-shift. It is reasonable because the Eq.~\eqref{eq:ie_total_loss} contains more than reconstruction loss.
Directly computing the anomaly region by comparing $x$ and $m$ may introduce extra differences in the normal region.
To handle this, we further introduce an Expert-Net as the second stage for high-fidelity reconstruction. The training and inference phases are as follow.

\noindent\textbf{Training phase.}
We first generate the impressions (we noted this set as $\MM$ below) for the training images, which is based on the training set for IE-Net. 
The anomaly-free images $\DD$ and their corresponding impressions $\MM$ construct the dataset, which is used for training the Expert-Net.
As illustrated in Fig.~\ref{fig:expert_pix2pix}, the Expert-Net is constructed by a detail extractor $E_S$, two encoders $\{E_X^A, E_X^B\}$, two decoders $\{D_X^A, D_X^B\}$.
Then Expert-Net aims to learn the invertible mapping between $\MM$ and $\DD$ through supervised learning. 

\noindent\textbf{Inference phase.}
When the Expert-Net training is complete, we generate the high-fidelity reconstruction $\hat{x}$ through $D_X^A(E_X^A(m), E_S(x))$. We can also get the na\"ive impression through $D_X^B(E_X^B(x))$. 

More specifically, $E_S$ encodes the detail vector $s$ of the input image $x$. Then $s$ is used for rectifying the generated details on $\hat{x}$. To make $E_S$ focus on local details, we use a small CNN with fewer layers than $E_X$ in implementation.
Inspired by the unsupervised image-to-image translation methods~\cite{R:huang2017arbitrary,R:huang2018munit} that use 
Adaptive Instance Normalization (AdaIN) layers to fuse different features for image generation, we also use AdaIN in Expert-Net. The AdaIN is defined as below
\begin{equation} \label{eq:AdaIN}
\text{AdaIN}(k, \gamma, \beta) = \gamma(\frac{k - f(k)}{g(k)}) + \beta,
\end{equation}
where $k$ is the activation of the previous convolutional layer, $f(\centerdot)$ and $g(\centerdot)$ are channel-wise mean and standard deviation. Take detail vector $s$ as input, $\gamma$ and $\beta$ are parameters generated by MLP.
The details of generated image $\hat{x}$ are also supervised by making its details vector $\hat{s}$ identical to the $s$. 

Motivated by the asynchronous learning methods~\cite{R:ge2020MMT, R:he2019moco}, we also enable the Expert-Net to mimic the procedure of impression extraction, but learn to map the $\DD$ to $\MM$ directly. We call the mimic version of impression $\hat{m}$ as na\"ve impression. The appearance of $\hat{m}$ may close to $m$ (\eg, blurry), however, it is not ensured that $\hat{m}$ is anomaly-free. Therefore, the difference between $m$ and $\hat{m}$ can also be a reference for anomaly detection. 
During training, the generated image $\hat{x}, \hat{m}$ and intermediate $\hat{s}$ are constrained to $x, m, s$ through L1 loss, respectively. 

\subsection{Perceptual measurement for anomaly detection} \label{sec:PM}

Traditional methods~\cite{A:akccay2019skip,A:schlegl2019fastAnnoGAN} identify the anomaly by using the pixel-wise difference (\eg, L1, MSE). 
To effectively detect the anomaly regions, we find that using perceptual measurement (PM) among images on anomaly detection can achieve better performance. Perceptual distance has also been successfully applied to other tasks such as image synthesis and style transfer~\cite{R:huang2018munit}. 
When inference, given an unknown image $x$, the corresponding $\{m, \hat{x}, \hat{m}\}$ are estimated through \proposed. 
The anomaly map $e$ of the unknown image is calculated through perceptual distance:
\begin{equation} \label{eq:anomaly_map}
\begin{aligned}
    e &= \Phi(x, \hat{x}, m, \hat{m}) \\
      &= \sum_l\lambda^e_l(\phi_l(x, \hat{x}) + \phi_l(m, \hat{m}) + \phi_l(x, m)),
\end{aligned}
\end{equation}
where $\phi(\centerdot)$ is the feature distance (\ie, perceptual distance, which measures the $L1$ distance of features in a pre-trained VGG-19 network).  

Based on the anomaly error map $e$, the anomaly region segmentation $y$ is calculated below
\begin{equation} \label{eq:anomaly_seg}
\begin{aligned}
    y(i, j) = \begin{cases}
1,  & \text{if } e(i, j) > \alpha; \\
0,  & \text{otherwise,}
\end{cases} 
\end{aligned}
\end{equation}
where $i, j$ are pixel index of error map $e$ and $\alpha$ is threshold for anomaly segmentation.

\begin{figure*} [htbp]
	\begin{center}
		\includegraphics[width=0.92\linewidth]{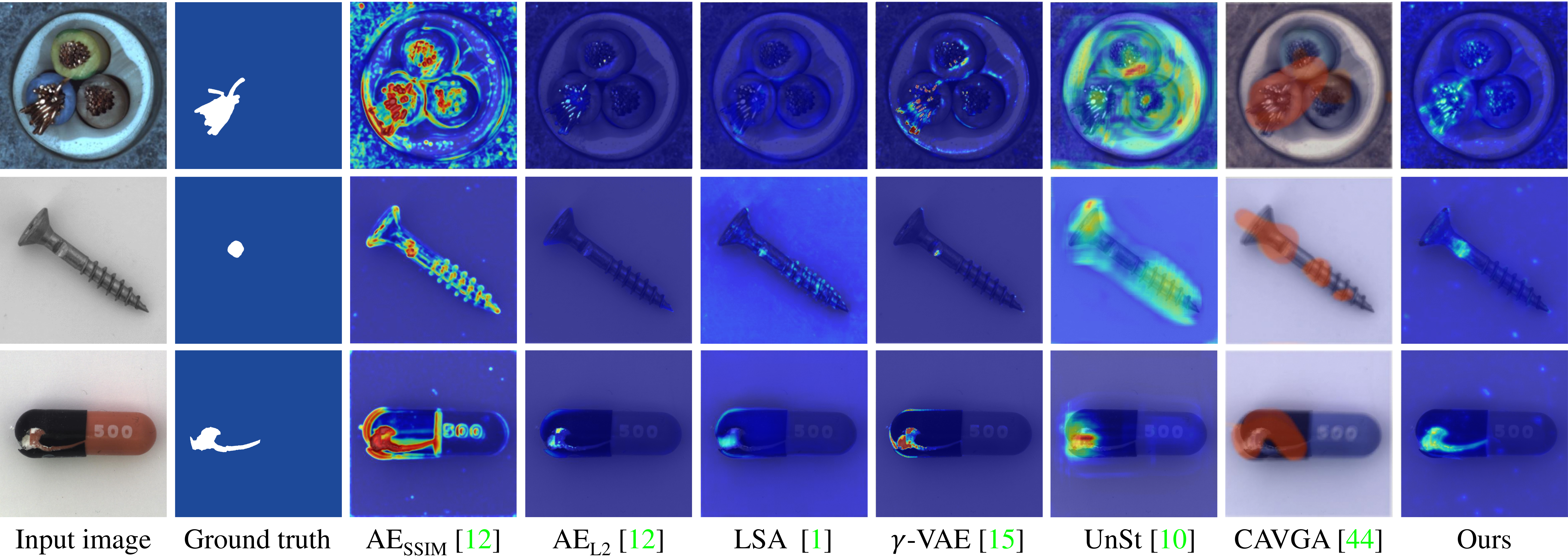} 
	\end{center}
	\vspace{-0.2cm}
	\caption{Qualitative comparisons among different methods on MVTec AD dataset.}
	\label{fig:cmp}
	\vspace{-0.2cm}
\end{figure*}

\begin{figure} [htbp]
	\begin{center}
		\includegraphics[width=\linewidth]{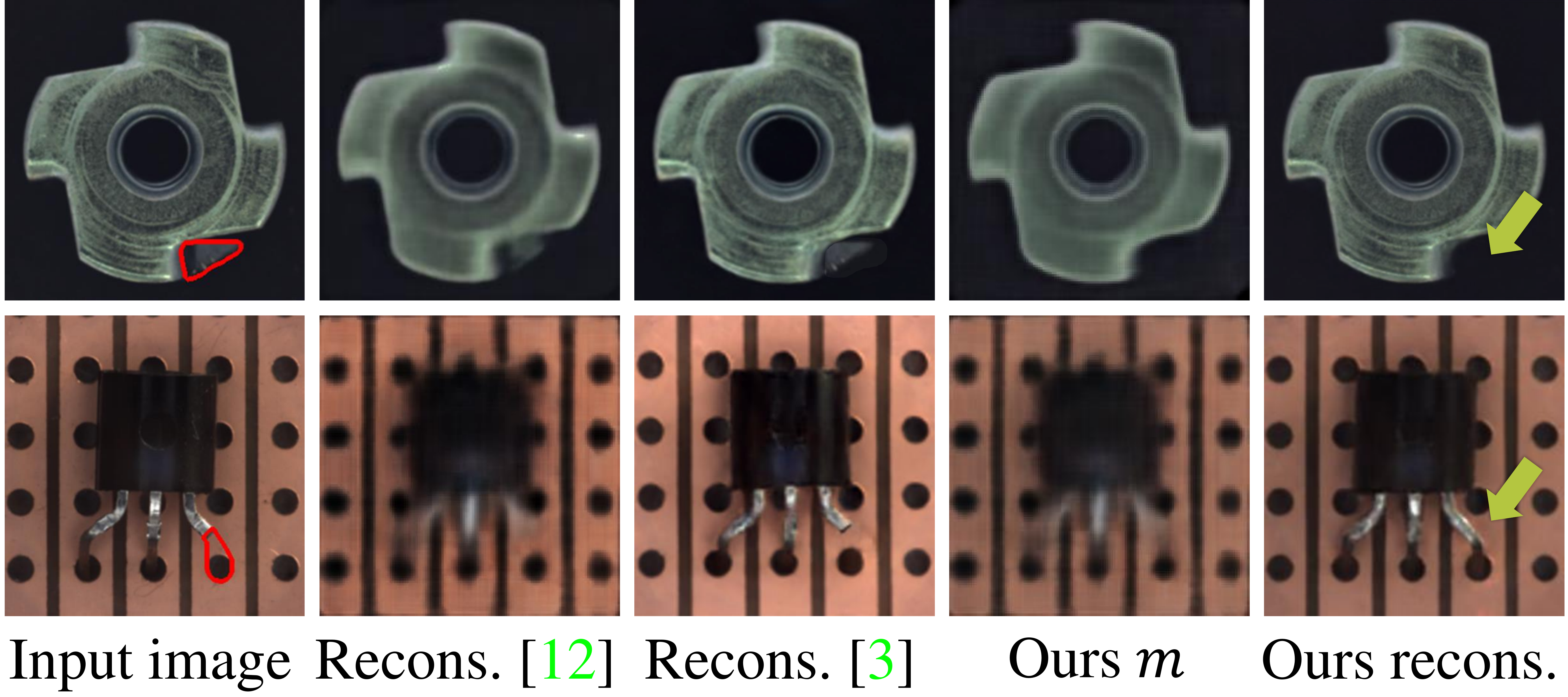} 
	\end{center}
	\vspace{-0.2cm}
	\caption{Our method generates \textbf{anomaly-free} impression $m$ and its \textbf{high-fidelity} reconstruction. The anomaly areas are highlighted in the input images with red contours.}
	\label{fig:cmp_recons}
	\vspace{-0.2cm}
\end{figure}

\begin{table*}[htbp]
	\begin{center}
		\setlength{\tabcolsep}{0.86mm}
		\renewcommand\arraystretch{1.2}
		\caption{Quantitative comparison of anomaly detection in category-specific IoU, mean IoU ($\overline{\text{IoU}}$) and mean AuROC ($\overline{\text{AuROC}}$) on the MVTec AD dataset. The best results are highlighted with bold font, and the second bests are underlined.}
		\label{tab:numerical_comparisons_sota}
		\small
		\begin{tabular}{lc cc cc cc cc cc}
			\toprule[1.2pt]
			
			Category		 & AnoGAN~\cite{A:schlegl2017unsupervised} & AVID~\cite{A:sabokrou2018avid} & $\text{AE}_\text{L2}$~\cite{A:bergmann2018improving} & $\text{AE}_\text{SSIM}$~\cite{A:bergmann2018improving} & LSA~\cite{A:abati2019LSA} & AD-VAE~\cite{A:liu2020towards} & $\gamma$-VAE~\cite{A:dehaene2020iterative} & UnSt~\cite{A:bergmann2020uninformed} & CAVGA~\cite{A:venkataramanan2020attention} & Ours \\
			\hline
			\specialrule{0em}{1pt}{1pt}
			
			Bottle     		 & 0.05 & 0.28 & 0.22 & 0.15 & 0.27 & 0.27 & 0.27 & 0.28 & \underline{0.34} & \textbf{0.37}  \\
			\rowcolor{mygray}
			Cable      		 & 0.01 & 0.27 & 0.05 & 0.01 & 0.36 & 0.18 & 0.26 & 0.11 & \textbf{0.38}    & \textbf{0.38} \\
			Capsule    		 & 0.04 & 0.21 & 0.11 & 0.09 & 0.22 & 0.11 & 0.24 & 0.24 & \underline{0.31} & \textbf{0.41} \\
			\rowcolor{mygray}
			Carpet     		 & 0.34 & 0.25 & 0.38 & 0.69 & 0.76 & 0.10 & \textbf{0.79} & 0.50 & 0.73 & \textbf{0.79} \\
			Grid       		 & 0.04 & 0.51 & 0.83 & \underline{0.88} & 0.20 & 0.02 & 0.36 & 0.19 & 0.38 & \textbf{0.89} \\
			\rowcolor{mygray}
			Hazelnut   		 & 0.02 & 0.54 & 0.41 & 0.00 & 0.41 & 0.44 & \underline{0.63} & 0.36 & 0.51 & \textbf{0.65} \\
			Leather    		 & 0.34 & 0.32 & 0.67 & 0.34 & 0.77 & 0.24 & 0.41 & 0.44 & \textbf{0.79} & \textbf{0.79} \\
			\rowcolor{mygray}
			Metal Nut  		 & 0.00 & 0.05 & 0.26 & 0.01 & 0.38 & \textbf{0.49} & 0.22 & 0.31 & 0.45 & \underline{0.47} \\
			Pill       		 & 0.17 & 0.11 & 0.25 & 0.07 & 0.18 & 0.18 & \underline{0.48} & 0.23 & 0.40 & \textbf{0.49} \\
			\rowcolor{mygray}
			Screw      		 & 0.01 & 0.22 & 0.34 & 0.03 & 0.38 & 0.17 & 0.38 & 0.17 & \textbf{0.48} & \underline{0.44} \\
			Tile       		 & 0.08 & 0.09 & 0.23 & 0.04 & 0.32 & 0.23 & 0.38 & 0.22 & \underline{0.38} & \textbf{0.40} \\
			\rowcolor{mygray}
			Toothbrush 		 & 0.07 & 0.43 & 0.51 & 0.08 & 0.48 & 0.14 & 0.37 & 0.21 & \textbf{0.57} & \underline{0.53} \\
			Transistor 		 & 0.08 & 0.22 & 0.22 & 0.01 & 0.21 & 0.30 & \underline{0.44} & 0.15 & 0.35 & \textbf{0.47} \\
			\rowcolor{mygray}
			Wood       		 & 0.14 & 0.14 & 0.29 & 0.36 & 0.41 & 0.14 & 0.45 & 0.16 & \textbf{0.59} & \textbf{0.59} \\
			Zipper     		 & 0.01 & \underline{0.25} & 0.13 & 0.10 & 0.14 & 0.06 & 0.17 & 0.08 & 0.16 & \textbf{0.30} \\
			\hline
			\specialrule{0em}{1pt}{1pt}
			\rowcolor{mygray}
			$\overline{\text{IoU}}$   & 0.09 & 0.26 & 0.33 & 0.19 & 0.37 & 0.20 & 0.39 & 0.24 & \underline{0.47} & \textbf{0.53} \\
			$\overline{\text{AuROC}}$ & 0.74 & 0.78 & 0.82 & 0.87 & 0.79 & 0.86 & 0.86 & 0.87 & \underline{0.89} & \textbf{0.90} \\
			\bottomrule[1.2pt]
		\end{tabular}
	\end{center}
	\vspace{-0.5cm}
\end{table*}

\section{Experiments}

To demonstrate the effectiveness of the proposed \proposed, extensive evaluation and rigorous analysis of ablation studies on a number of datasets are performed.

\subsection{Experimental setup}

\noindent\textbf{Benchmark datasets.}
We evaluate our \proposed~ on the MVTec AD~\cite{A:bergmann2019mvtec} with 15 different objects and textures for anomaly detection. Furthermore, MNIST~\cite{B:lecun1998MNIST}, Fashion MNIST~\cite{B:xiao2017FashionMNIST}, and CIFAR-10~\cite{B:krizhevsky2009CIFAR10} are also used for anomaly detection.

\noindent\textbf{Baseline methods.}
We compare \proposed~ with OCGAN~\cite{A:perera2019ocgan},  AnoGAN~\cite{A:schlegl2017unsupervised},  AVID~\cite{A:sabokrou2018avid}, $\text{AE}_\text{L2}$~\cite{A:bergmann2018improving},  $\text{AE}_\text{SSIM}$~\cite{A:bergmann2018improving}, LSA~\cite{A:abati2019LSA}, and AD-VAE~\cite{A:liu2020towards} with their publicly official code. Since the official project of UnSt~\cite{A:bergmann2020uninformed} and $\gamma$-VAE~\cite{A:dehaene2020iterative} are not publicly available, we use their third-party implementations\footnote{\scriptsize{\url{https://github.com/denguir/student-teacher-anomaly-detection}}}\footnote{\scriptsize{\url{https://github.com/dbbbbm/energy-projection-anomaly}}}, respectively. We have also compared our method with the unsupervised version of CAVGA~\cite{A:venkataramanan2020attention}.

\noindent\textbf{Implementation details.}
The detailed implementation of the proposed method is organized as follow:
\begin{itemize}
	\item \textit{IE-Net}. The encoder $E_M$ is constructed by 4 inception blocks~\cite{B:szegedy2015InceptionNet}. Each of them is followed by a max-pooling layer for down-sampling the feature map. The decoder $D_M$ contains 4 inception blocks with a 2$\times$ nearest up-sample layer, then followed by a $1\times 1$ convolutional layer with sigmoid activation for the reconstruction of the impression. The MLP is constructed by a stack of three fully connected layers. The discriminator $T$ is constructed by a 4-layer MLP with a sigmoid activation layer at the endpoint.
	\item \textit{Expert-Net.} Based on MUNIT~\cite{R:huang2018munit}, the encoders ($E_X^A, E_X^B$) of the Expert-Net contains 4 convolutional blocks and 2 residual blocks. The decoder is constructed by 2 residual blocks with AdaIN as normalization layer, then followed 4 convolutional blocks with upsamples to reconstruct the image. The \textit{detail extractor} $E_S$ contains 3 convolutional layers and follows an adaptive average pooling layer.
	\item \textit{PM.} To effectively involve different features for anomaly detection, we select the layers `conv1\_2', `conv2\_2', and `conv3\_4' in the VGG-19 network and set $\lambda^e_1 = \lambda^e_2 = \lambda^e_3 = 1$ in our experiments. 
	We empirically set $\alpha=0.5$ for anomaly segmentation.
\end{itemize}

\begin{figure*} 
	\begin{center}
		\includegraphics[width=0.9\linewidth]{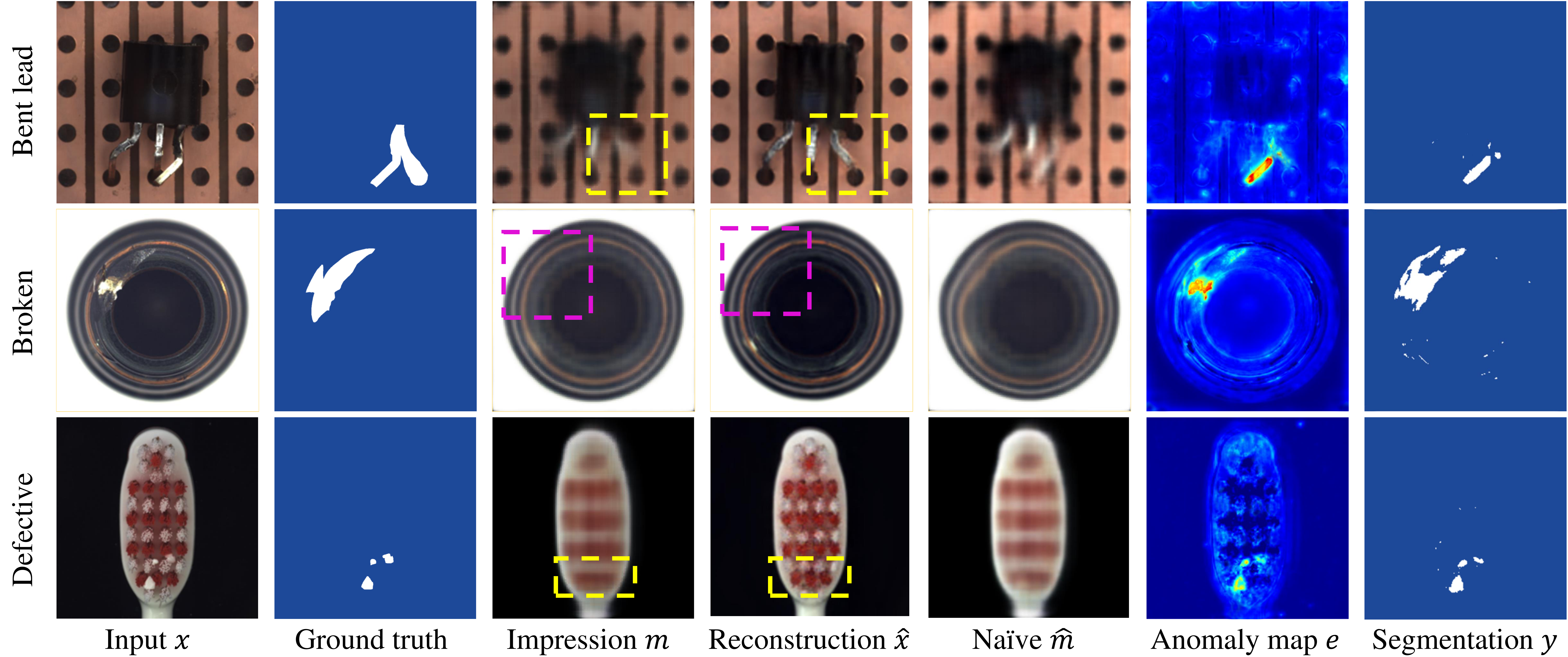} 
	\end{center}
	\vspace{-0.2cm}
	\caption{Interpretation of the intermediate results in \proposed. As illustrated in dashed boxes, the impression $m$ is robust to anomalous regions, while the na\"ive impression $\hat{m}$ is affected by the anomalous regions.
	}
	\vspace{-0.2cm}
	\label{fig:viz}
\end{figure*}

\subsection{Comparisons to SOTAs}
We evaluate our method on 15 category-specific dataset with high-resolution images (\ie, MVTec AD~\cite{A:bergmann2019mvtec}) and three low-resolution datasets (MNIST~\cite{B:lecun1998MNIST}, Fashion MNIST~\cite{B:xiao2017FashionMNIST}, and CIFAR-10~\cite{B:krizhevsky2009CIFAR10}).

\textbf{MVTec AD dataset.} 
For all experiments on MVTec AD, images are re-scaled to $w = h = \text{256}$ for training and inference. We train the IE-Net on anomaly-free images for 200 epochs with batch size 4. We use SGD with initial learning rate $\text{10}^{-3}$ and momentum 0.9. For Expert-Net, we use Adam with learning rate $\text{10}^{-3}$.
Fig.~\ref{fig:cmp} illustrates the visual comparisons among different methods. Our method localizes more accurate and fine-grained anomalous regions than the compared methods. Since the CAVGA is not publicly available, we directly use the results from their supplement materials. As illustrated in Fig.~\ref{fig:cmp_recons}, our method makes anomaly-free and high-fidelity reconstructions when compared to the two typical one-state methods~\cite{A:bergmann2018improving,A:akccay2019skip}.
Table~\ref{tab:numerical_comparisons_sota} shows the quantitative comparison over Intersection over Union (IoU) and Area under ROC curve (AuROC). Our method outperforms the state-of-the-art method (CAVGA) in mean IoU by 6\%. 
 
\textbf{MNIST, Fashion MNIST and CIFAR-10.}
On the MNIST, Fashion-MNIST, and CIFAR-10 datasets, we follow the same settings as in~\cite{A:deecke2018image} (\ie, training/testing uses a single class as normal and the rest of the classes as anomalous. Each image is zoomed to $w = h = \text{64}$ for training and inference. Because the resolution and input size of these three datasets are lower than those in MVTec AD dataset, we adjust the architecture of IE-Net and Expert-Net to fit these cases. Specifically, we reduce the number of inception blocks to 3 in IE-Net. Meanwhile, we set the number of residual blocks to 1 in Expert-Net.
Table~\ref{tab:numerical_ablation} shows the quantitative comparison results. We find the ablation studies verify different components of our framework. Our proposed outperforms the other methods for many settings.

\begin{table}[htbp]
	\begin{center}
		\setlength{\tabcolsep}{1.2mm}
		\renewcommand\arraystretch{1.2}
	    \vspace{-0.2cm}
		\caption{Quantitative comparison of different methods and ablation study on MNIST ($D_M$), Fashion MNIST ($D_F$) and CIFAR-10 ($D_C$) datasets. Here we use mean AuROC as the evaluation metric.}
		\small
		\label{tab:numerical_ablation}
		\begin{tabular}{lc cc cc cc}
			\toprule[1.2pt]
			
			Existing method		            &            &            &&            & $D_M$   & $D_F$   & $D_C$     \\
			\hline
			\specialrule{0em}{1pt}{1pt}
			OCGAN~\cite{A:perera2019ocgan} &            &            &&            & 0.975   &  0.895   & 0.657     \\
			\rowcolor{mygray}
			AnoGAN~\cite{A:schlegl2017unsupervised}               &&            &            &            & 0.937   &  0.824 & 0.612      \\
			SkipGA~\cite{A:akccay2019skip}               &&            &            &            & 0.941   &  0.807 & 0.731      \\
			\rowcolor{mygray}
			LSA~\cite{A:abati2019LSA}                              &&            &            &            & 0.975   &  0.641   & 0.876     \\
			$\text{AE}_\text{L2}$~\cite{A:bergmann2018improving}   &&            &            &            & 0.983   &  0.747   & 0.790      \\
			\rowcolor{mygray}
			CapsNet~\cite{A:li2020exploring}                       &&            &            &            & 0.871   &  0.679  & 0.531      \\
			UnSt~\cite{A:bergmann2020uninformed}                   &&            &            &            & 0.993   &  -         & 0.803      \\
			\rowcolor{mygray}
			CAVGA~\cite{A:venkataramanan2020attention}             &&            &            &            & 0.986   &  0.885  & 0.737      \\
			\hline
			\specialrule{0em}{1pt}{1pt}
			Ablation                & $L_{info}$ & \textit{EN}& Guide      & $\hat{m}$  &        &         &           \\
			\hline
			\specialrule{0em}{1pt}{1pt}
			Baseline                &            &            &            &            & 0.979  &  0.743  & 0.787     \\
			\rowcolor{mygray}
			Ours                    & \checkmark &            &            &            & 0.981  &  0.751  & 0.803     \\
			Ours                    &            & \checkmark &            &            & 0.989  &  0.885  & 0.861      \\
			\rowcolor{mygray}
			Ours                    & \checkmark & \checkmark &            &            & 0.988  &  0.890  & 0.872      \\
			Ours                    &            & \checkmark & \checkmark &            & 0.993  &  0.891  & 0.876      \\
			\rowcolor{mygray}
			Ours                    & \checkmark & \checkmark & \checkmark &            & 0.992 &  0.894  & 0.881     \\
			Ours                    & \checkmark & \checkmark & \checkmark & \checkmark & \textbf{0.994}  &  \textbf{0.899}  & \textbf{0.884}     \\
			\bottomrule[1.2pt]
		\end{tabular}
	\end{center}
	\vspace{-0.8cm}
\end{table}

\begin{figure} [htbp]
	\begin{center}
		\includegraphics[width=0.9\linewidth]{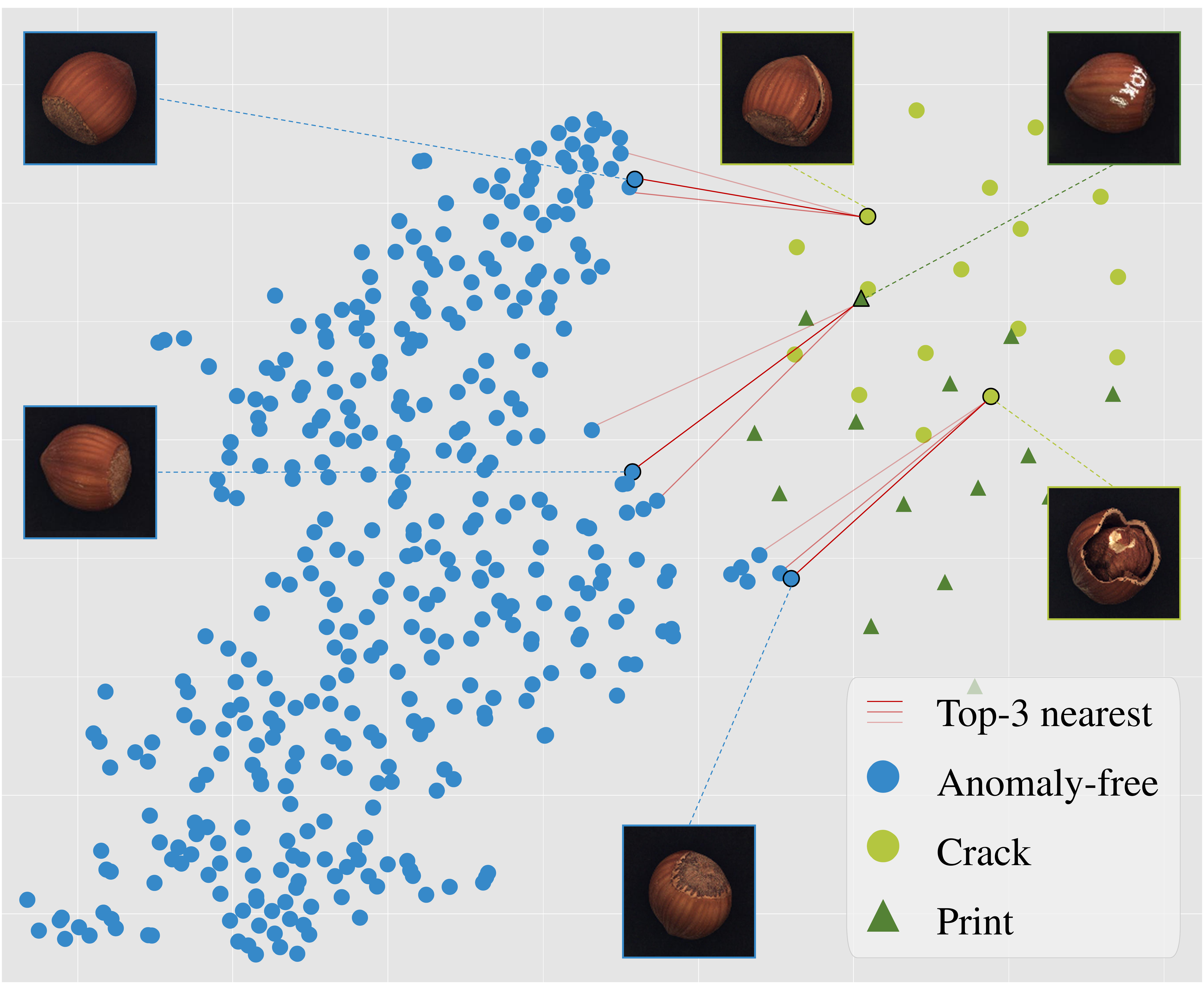} 
	\end{center}
	\vspace{-0.2cm}
	\caption{Analysis of the learned features in IE-Net. Feature distributions of anomaly-free images (blue) and anomalous images (green) are illustrated via t-SNE~\cite{B:maaten2008visualizing}. }
	\label{fig:analysis}
	\vspace{-0.4cm}
\end{figure}

\begin{figure*} 
	\begin{center}
		\includegraphics[width=0.9\linewidth]{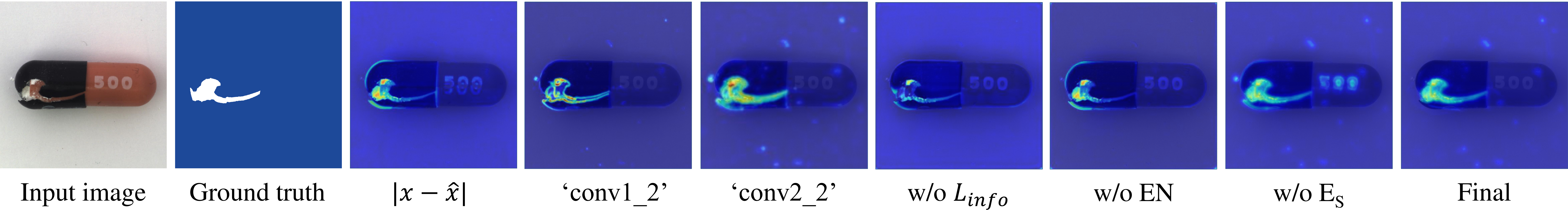} 
	\end{center}
	\vspace{-0.2cm}
	\caption{Qualitative comparison of different combinations of our method on anomaly detection.} 
	\label{fig:cmp_ablation}
	\vspace{-0.4cm}
\end{figure*}

\begin{figure} [htbp]
	\begin{center}
		\includegraphics[width=\linewidth]{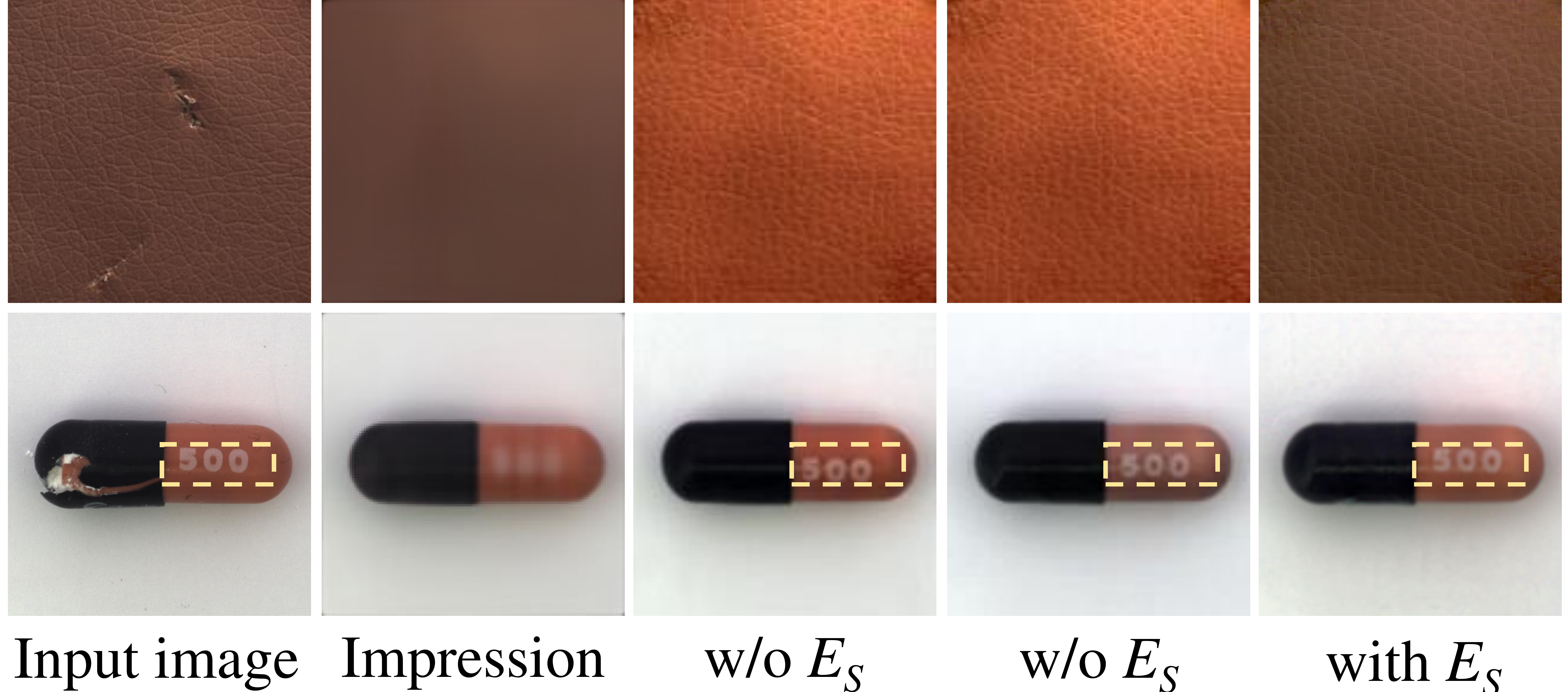} 
	\end{center}
	\vspace{-0.2cm}
	\caption{Illustration of the effectiveness of detail guidance module $E_S$. Two results are provided without $E_S$ (Column 3-4). With $E_S$, some inconsistencies in details \eg, color-shift (top row) and misplaced text (\ie, `500', dashed box in bottom row) are tackled.}
	\label{fig:cmp_detail}
	\vspace{-0.2cm}
\end{figure}

\subsection{Interpretation and analysis}

\noindent\textbf{Interpretation of \proposed.}

Interpretation can be performed with two folds: illustration of intermediate results of \proposed~ and visualization of feature distributions.
As illustrated in Fig.~\ref{fig:viz}, it is clear that the impression $m$ automatically fixes the anomaly region. Based on the impression, the Expert-Net reconstructs the exact details of the anomaly-free image $\hat{x}$. Meanwhile, the na\"ive impression version $\hat{m}$ often contains the anomaly area with the total image become blurry. Next, the anomaly region $e$ is calculated by a pixel-wise scoring as \eq~\eqref{eq:anomaly_map}, in which pixels in the anomaly region have higher scores than that in the anomaly-free region. 
Fig.~\ref{fig:analysis} shows the learned distributions of anomaly-free images and anomalous images. It can be found that the anomaly image and anomalous image are with different mean and variance. Furthermore, based on the mutual information, one can be found the k-Nearest anomaly-free images when the anomalous image is given. For instance, the cracked hazelnut share a similar pose with the matched anomaly-free one at the top of Fig.~\ref{fig:analysis}. Here two types of anomalous images are given (\ie, crack and print).

\noindent\textbf{Ablation study.}

To evaluate how different parts of \proposed~contribute to the final performance on the two tasks, we conduct rigorous ablation studies by removing or replacing a subset of models. Details of different baselines are described as follows:
	
	\begin{itemize}
		\setlength{\itemsep}{2pt}
		\setlength{\parsep}{-2pt}
		\setlength{\parskip}{0pt}
		
		\item \emph{Without maximizing mutual information}~(Without $L_{info}$). To illustrate the contribution of the mutual information learning for impression extraction, we remove this part and re-train the remaining model.
		\item \emph{Effectiveness of Expert-Net}~(Without \textit{EN}). To examine the effectiveness of Expert-Net in anomaly detection, we remove Expert-Net totally and the anomaly error map $e$ is calculated by $e = \sum_l(\phi_l(x, m))$.
		\item \emph{Detail Guidance module}~(Without $E_S$). To examine the contribution of the guidance from detailed information, we remove the $E_S$ and feed the input of AdaIN in the decoder of Expert-Net with random weights of Gaussian distributions (\ie, $N(0, 1)$).
	\end{itemize}
The ablation studies are mainly performed on 10 categories on the MNIST, Fashion MNIST, and CIFAR-10 datasets. The mean AuROC of each dataset is shown in Table~\ref{tab:numerical_ablation}. Since for UnSt~\cite{A:bergmann2020uninformed}, one teacher net and 5 student nets are trained, it performs exceptionally well on these small datasets. However, on average, our method still outperforms all evaluated approaches.
Furthermore, the visual comparison is provided in the 6th - 8th column in Fig~\ref{fig:cmp_ablation}. Without $L_{info}$, the anomaly region cannot be corrected detected. Without \textit{EN}, misaligned edges occur in anomaly detection. Anomaly-free regions on the capsule are mistakenly detected as anomaly when the $E_S$ is removed (the detail code $s$ is generated with random numbers). More visual results are provided in Fig.~\ref{fig:cmp_detail}.

\noindent\textbf{Different measurements for anomaly detection.}

A variety of measurements for anomaly detection has been tested on MVTec AD dataset. Numerical comparisons of different measurements on 4 objects are listed in Table~\ref{tab:numerical_ablation_percep}. The combination of `conv1\_2', `conv2\_2' and `conv3\_4' (\ie, \textit{PM}) gets the best performance on different objects.
We find the difference of features from  `conv2\_2' gets secondary performance. The performance is deceased in `conv3\_4' since the resolution is too small. 
As illustrated in Fig.~\ref{fig:cmp_ablation}, 
It can be clearly found that the pixel-level or low-level differences like $|x - \hat{x}|$ and `conv1\_2' are sensible on the local difference, but often misled by misalignment. Meanwhile, the high-level difference (\ie, `conv2\_2') gets relative correct regions. The result of \textit{PM} gets the best visual quality.

\begin{table}[htbp]
	\begin{center}
		\setlength{\tabcolsep}{3mm}
		\renewcommand\arraystretch{1.0}
		\caption{Ablation study. Performance comparison among differences between input and intermediate results and different choices of layers in VGG-19 on part of MVTec AD dataset.}
		\small
		\label{tab:numerical_ablation_percep}
		\begin{tabular}{lc cc c}
			\toprule[1.2pt]
			
			Measurement     & Bottle & Cable & Capsule & Hazelnut  \\
			\hline
			\specialrule{0em}{1pt}{1pt}
			$|x - \hat{x}|$ & 0.23   & 0.17  & 0.33    & 0.39      \\
			\rowcolor{mygray}
			$|x - m|$       & 0.29   & 0.20  & 0.21    & 0.47      \\
			$|m - \hat{m}|$ & 0.15   & 0.06  & 0.15    & 0.41      \\
			\rowcolor{mygray}
			`conv1\_2'      & 0.30   & 0.11  & 0.33    & 0.39      \\
			`conv2\_2'      & 0.33   & 0.31  & 0.35    & 0.54      \\
			\rowcolor{mygray}
			`conv3\_4'      & 0.24   & 0.20  & 0.31    & 0.49      \\
			\textit{PM}     & \textbf{0.37} & \textbf{0.38} & \textbf{0.41} & \textbf{0.65}  \\
			\bottomrule[1.2pt]
		\end{tabular}
	\end{center}
	\vspace{-0.5cm}
\end{table}

\noindent\textbf{Effect of the model capacity.}

Finally, we investigate the relationship between the model capacity and anomaly detection performance. Numerical results on MVTecAD Hazelnut are reported in Table~\ref{tab:param_analysis}. We find that more learnable parameters, which result in better reconstruction ability of an AutoEncoder (AE)~\cite{A:bergmann2018improving}, do not help the anomaly detection. This confirms our observation in Fig.~\ref{fig:diff}. On the other hand, benefiting from the proposed two-stage reconstruction scheme, our method outperforms UnSt~\cite{A:bergmann2020uninformed} with a similar number of parameters.

\begin{table}[htbp]
	\renewcommand\arraystretch{1.0}
    \setlength{\tabcolsep}{1.5mm}
    \centering
    \small
    \caption{Effect of parameters' number in training.}
	\label{tab:param_analysis}
    \begin{tabular}{lccccc}
		\toprule[1.2pt]
         Methods      &  AE-64 & AE-128 & AE-256 & UnSt~\cite{A:bergmann2020uninformed} & Ours  \\
         \hline
         \# Param (M)        &  2.98  & 11.89 &  47.47  & 14.28 & 16.77 \\
		 \rowcolor{mygray}
         IoU ($\uparrow$)    &  0.301 & 0.188 & 0.197   & 0.363 & \textbf{0.649} \\
         AuROC ($\uparrow$)  &  0.896 & 0.954 & 0.945   & 0.937 & \textbf{0.976} \\
		\bottomrule[1.2pt]
    \end{tabular}  
\vspace{-0.3cm}
\end{table}

\section{Conclusion}

In this paper, we propose a novel method (\ie, the Unsupervised Two-stage Anomaly Detection) for unsupervised anomaly detection in natural images. In particular, we propose to utilize a two-stage framework (\ie, IE-Net, Expert-Net) for anomaly detection. The IE-Net and Expert-Net are used to generate high-fidelity and anomaly-free reconstructions of the input. \proposed~ generates rich and intuitive intermediate results, make the framework interpretable. Extensive experiments demonstrate state-of-the-art performance on different datasets, which contain different types of real-world objects and textures.

{\small
\bibliographystyle{ieee_fullname}
\bibliography{egbib}
}

\appendix

\end{document}